\def\year{2020}\relax
\documentclass[letterpaper]{article} 
\usepackage{aaai20}  
\usepackage{times}  
\usepackage{helvet} 
\usepackage{courier}  
\usepackage[hyphens]{url}  
\usepackage{bm}
\usepackage{graphicx} 
\usepackage{amsmath}
\usepackage{multirow}
\usepackage{graphicx}  
\title{Deep Object Co-segmentation via Spatial-Semantic Network Modulation}
\author{Kaihua Zhang\textsuperscript{\rm 1}, Jin Chen\textsuperscript{\rm 1}, Bo Liu\textsuperscript{\rm 2}\thanks{Corresponding author}, Qingshan Liu\textsuperscript{\rm 1}
\\
\textsuperscript{\rm 1}B-DAT and CICAEET, Nanjing University of Information Science and Technology, Nanjing, China
\\
\textsuperscript{\rm 2}JD Finance America Corporation
\\
\{zhkhua, kfliubo\}@gmail.com 
}
\begin{document}
\maketitle

\begin{abstract}
Object co-segmentation is to segment the shared objects in multiple relevant images, which has numerous applications in computer vision.
This paper presents a spatial and semantic modulated deep network framework for object co-segmentation. A backbone network is adopted to extract multi-resolution image features. With the multi-resolution features of the relevant images as input, we design a spatial modulator to learn a mask for each image. The spatial modulator captures the correlations of image feature descriptors via unsupervised learning. The learned mask can roughly localize the shared foreground object while suppressing the background. For the semantic modulator, we model it as a supervised image classification task. We propose a hierarchical second-order pooling module to transform the image features for classification use. The outputs of the two modulators manipulate the multi-resolution features by a shift-and-scale operation so that the features focus on segmenting co-object regions. The proposed model is trained end-to-end without any intricate post-processing. Extensive experiments on four image co-segmentation benchmark datasets demonstrate the superior accuracy of the proposed method compared to state-of-the-art methods.
The codes are available at \url{http://kaihuazhang.net/}.
%
\end{abstract}
\section{Introduction}
\hspace*{1em}
As a special case of image object segmentation, object co-segmentation refers to the task of jointly discovering and segmenting the objects shared in a group of images.
It has been widely used to support various computer vision applications, such as interactive image segmentation~\cite{kamranian2018iterative}, 3D reconstruction~\cite{mustafa2017semantically} and object co-localization~\cite{Wei2017Unsupervised,Han2018Robust}, to name a few.

%
%
Image features that characterize the co-objects in the image group are vital for a co-segmentation task.
Conventional approaches use the hand-crafted cues such as color histograms, Gabor filter outputs and SIFT descriptors as feature representations~\cite{yuan2014novel,dai2013cosegmentation,lee2015multiple}.
Those hand-crafted features cannot well handle the challenging cases in co-segmentation such as background clutter and large-scale appearance variations of the co-objects in images.
In recent years, deep-learning-based co-segmentation methods have attracted much attention. For example,~\cite{li2018deep,chen2018semantic} leverage a Siamese network architecture for object co-segmentation and an attention mechanism is used to enhance the co-object feature representations. These methods have shown superior performance compared to the traditional methods~\cite{yuan2014novel,dai2013cosegmentation,lee2015multiple}, which inspire us to explore a deep-learning-based solution to object co-segmentation.

\begin{figure}[t]
\centering
\includegraphics[width=0.9\columnwidth]{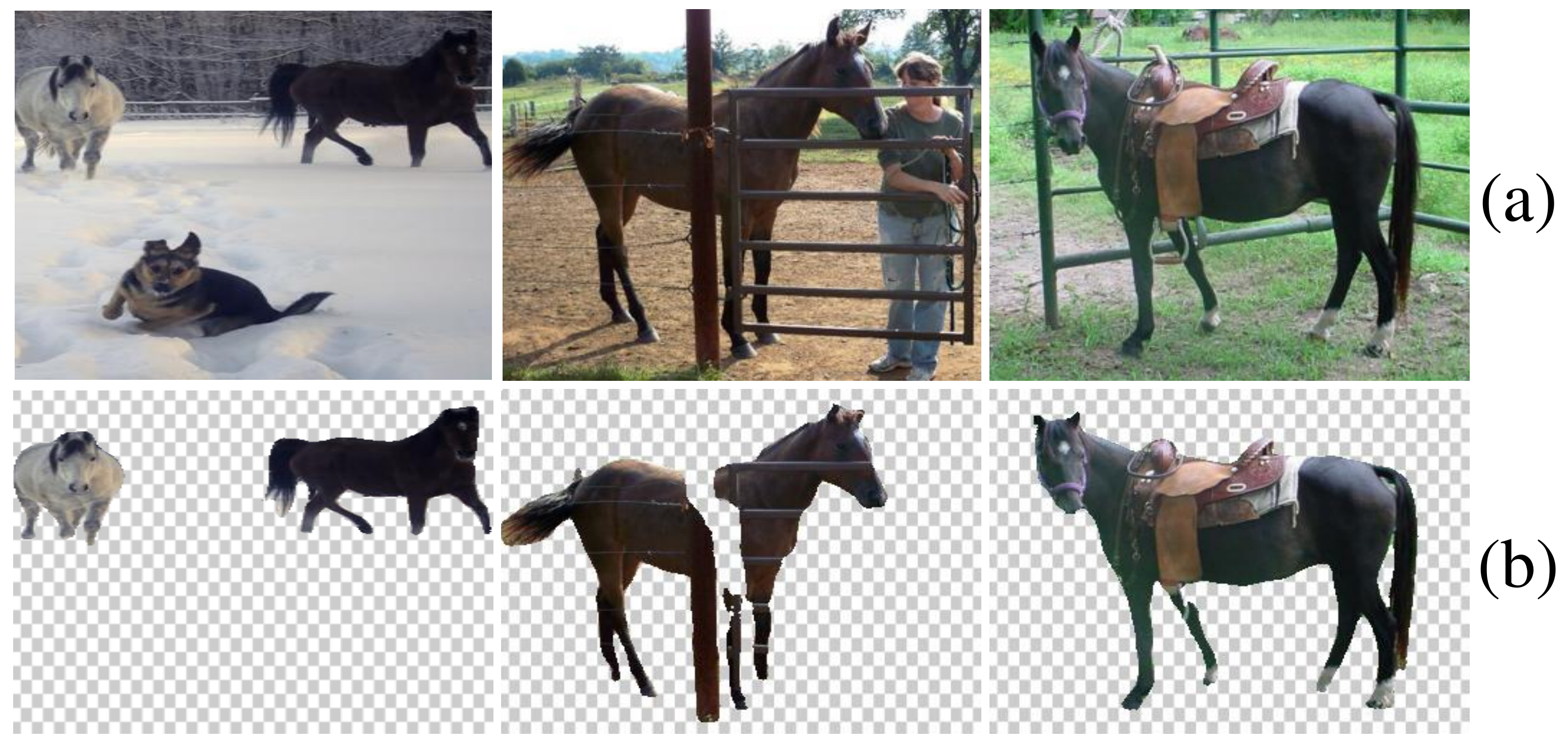} 
\caption{Object co-segmentation examples by our approach. (a) Horse group; (b) Horse group co-segmentation maps.}
\label{colocosegexample}
\end{figure}
One critical property of object co-segmentation is that the co-objects in images belong to the same semantic category. Those co-objects usually occupy part of each image. One illustrative example is shown in Figure~\ref{colocosegexample}. It is desirable that the deep convolutional network layers, being used as a feature extractor, are targeted on modelling the co-objects. To this end, we propose a spatial-semantic modulated network structure to model this property. The two modulators are achieved by the designed group-wise mask learning branch and co-category classification branch, respectively. We summarize the technical contributions of this work as follows:
\begin{quote}
\begin{itemize}
\item
We propose a spatial-semantic modulated deep network for object co-segmentation. Image features extracted by a backbone network are used to learn a spatial modulator and a semantic modulator. The outputs of the modulators guide the image features up-sampling to generate the co-segmentation results. The network parameter learning is formulated into a multi-task learning task, and the whole network is trained in an end-to-end manner.
\item
For the spatial modulation branch, an unsupervised learning method is proposed to learn a mask for each image. With the fused multi-resolution image features as input, we formulate the mask learning as an integer programming problem. Its continuous relaxation has a closed-form solution. The learned parameter indicates whether the corresponding image pixel corresponds to foreground or background.
\item
In the semantic modulation branch, we design a hierarchical second-order pooling (HSP) operator to transform the convolutional features for object classification. Spatial pooling (SP) is shown to be able to capture the high-order feature statistical dependency~\cite{gao2019global}. The proposed HSP module has a stack of two SP layers. They are dedicated to capturing the long-range channel-wise dependency of the holistic feature representation. The output of the HSP layer is fed into a fully-connected layer for object classification and used as the semantic modulator.
\end{itemize}
\end{quote}

We conduct extensive evaluations on four object co-segmentation benchmark datasets~\cite{faktor2013co,rubinstein2013unsupervised}, including the sub-set of MSRC, Internet, the sub-set of iCoseg and PASCAL-VOC datasets. The proposed model achieves a significantly higher accuracy than state-of-the-art methods. Especially, on the most challenging PASCAL-VOC dataset, our method outperforms the second best-performing state-of-the-art approach~\cite{li2018deep} by $6\%$ in terms of average Jaccard index $\mathcal{J}$.
The rest of this work is organized as follows: In~\S~\ref{sec:relatedwork}, we introduce the related works of
our study. \S~\ref{sec:proposedapproach} describes the proposed framework and its main components. Afterwards, comprehensive experimental evaluations are presented in~\S~\ref{sec:experiments}. Finally, we conclude this work in~\S~\ref{sec:conclusions}.
\section{Related Work}
\label{sec:relatedwork}
\subsection{Object Co-segmentation}
\hspace*{1em}
A more comprehensive literature review about image co-segmentation can be found in~\cite{zhu2016beyond}. Existing object co-segmentation methods can be roughly grouped into four categories including graph-based model, saliency-based model, joint processing model and deep learning model.
Conventional approaches such as \cite{yuan2014novel,collins2012random,lee2015multiple} assume the pixels or superpixels in the co-objects can be grouped together and then they formulate co-segmentation as a clustering task to search for the co-objects.
Saliency-detection-based methods assume regions of interest in the images are usually the co-objects to be segmented. They conduct image co-segmentation through detecting the regions that attract human attention most. Representative models include~\cite{tsai2018image,zhang2019co,lu2019survey}.
The work in~\cite{dai2013cosegmentation,jerripothula2017object} employs a coupled framework for co-skeletonization and co-segmentation tasks so that they are well informed by each other, and benefit each other synergistically. The idea of joint processing can exploit the inherent interdependencies of two tasks to achieve better results jointly.
%
%
Recently, \cite{li2018deep,chen2018semantic} respectively propose an end-to-end deep-learning-based method for object co-segmentation using a Siamese encoder-decoder architecture and a semantic-aware attention mechanism.
\subsection{Network Modulation}
\hspace*{1em}
Modulation module has been proved to be an effective way to manipulate network parameter learning. The modulator can be modelled as parameters or output of the auxiliary branch that are used to guide the main branch parameter learning. In the segmentation method~\cite{dai2015convolutional}, an image mask is used as a modulator for background removal. In the Mask R-CNN model~\cite{he2017mask}, a classification branch is used to guide the segmentation branch learning. Feature-wise linear modulation is a widely-used scheme, which has been applied to object detection~\cite{Lin2017Feature} and graph neural networks learning~\cite{brockschmidt}. In visual reasoning problem, network modulation is used to encode the language information~\cite{de2017modulating,perez2018film}. The attention module in the image caption model~\cite{chen2017sca} can be viewed as a modulator. \cite{yang2018efficient} proposes to model the visual and spatial information by a modulator for video object segmentation. In~\cite{flores2019saliency}, a saliency detection branch is added to an existing CNN architecture as a modulator for fine-grained object recognition. A cross-modulation mechanism is proposed in~\cite{prol2018cross} for few-shot learning.
\begin{figure*}[t]
\centering
\includegraphics[width=1\textwidth]{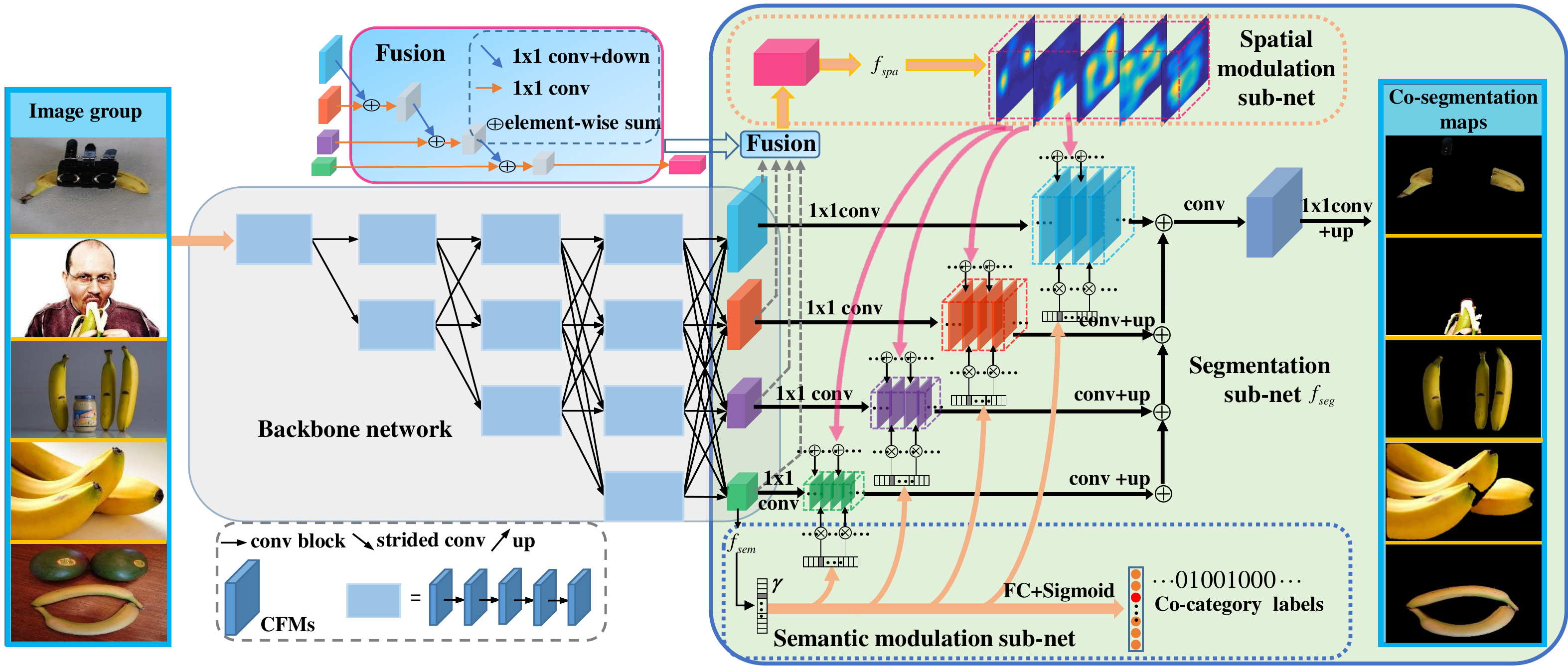} 
\caption{Overview of the proposed object co-segmentation framework. Firstly, a group of images are fed into the backbone network to yield a set of muti-resolution CFMs. Then, the CFMs are modulated by a group of spatial heatmaps and a feature channel selector vector. The former is generated by a clustering approach that can well capture the coarse localizations of the co-objects in the images. Under the supervision of co-category labels, the latter is obtained by learning a group-wise semantic representation that indicates the importance of the feature channels. Finally, the multi-resolution modulated CFMs are fused in a way similar to the feature pyramid network (FPN)~\cite{Lin2017Feature} to produce the co-segmentation maps. `\textbf{conv}', `\textbf{FC}', `\textbf{up}' and `\textbf{down}' are short for convolutional, fully-connected, upsampling and downsampling layers, respectively.}
\label{network}
\end{figure*}
\section{Proposed Approach}
\label{sec:proposedapproach}
\subsection{Problem Formulation}
\hspace*{1em}
Figure~\ref{network} presents an overview of our model. Given a group of $N$ images $\mathcal{I}=\{I^n\}_{n=1}^N$ containing co-objects of a specific category, our objective is to learn a feed-forward network $f$ that produces a set of object co-segmentation masks $\mathcal{M}=\{\textit{\textbf{M}}^n\}_{n=1}^N$:
\begin{equation}
\mathcal{M} = f(\mathcal{I};\bm{\theta}),
\end{equation}
where $\bm{\theta}$ denotes the network parameters to be optimized.
The network $f$ is composed of three sub-networks: spatial modulation sub-net $f_{spa}$, semantic modulation sub-net $f_{sem}$ and segmentation sub-net $f_{seg}$.
The renowned SPP-Net~\cite{he2015spatial} has shown that the convolutional feature maps (CFMs) for object recognition encode both spatial layouts of objects (by their positions) and the semantics (by strengths of their activations).
Inspired by this model, we design $f_{spa}$ and $f_{sem}$ to encode the spatial and semantic information of the co-objects in $\mathcal{I}$, respectively.
The two modulators guide the convolution layers learning in $f_{seg}$ to focus on the co-objects in the images.
Specifically, the sub-net $f_{spa}$ is to learn a mask for each image to coarsely localize the co-object in it.
Given the input CFMs $\{{\bm\varphi}(I^n)\}_{n=1}^N$ produced by fusing all the output CFMs of our backbone network, the sub-net $f_{spa}$ produces
a set of spatial masks $\mathcal{S}=\{\textit{\textbf{S}}^n\in \Re^{w\times h}\}_{n=1}^N$ with width $w$ and height $h$:
\begin{equation}
\label{eq:spa modulator}
\mathcal{S}=f_{spa}(\{{\bm\varphi}(I^n)\}_{n=1}^N;\bm{\theta}_{spa}),
\end{equation}
where $\bm{\theta}_{spa}$ denotes the corresponding network parameters to be optimized.
Although the coarse spatial layout information of the co-objects in all images can be embedded into $\mathcal{S}$ in (\ref{eq:spa modulator}), the useful high-level semantic information that are essential to differentiate co-objects from distractors fails to be transferred into $\mathcal{S}$.
To address this issue, we further propose $f_{sem}$ as a complement.
The sub-net $f_{sem}$ learns a channel selector vector $\bm{\gamma}\in \Re^d$ with $d$ channels. The entries of $\bm{\gamma}$ indicate the importance of feature channels, that is
\begin{equation}
\label{eq:semantic modulator}
\bm{\gamma}=f_{sem}(\{{\bm\phi}(I^n)\}_{n=1}^N;\bm{\theta}_{sem}),
\end{equation}
where $\bm{\phi}$ denotes the output CFMs with the lowest resolution generated by our backbone network, and $\bm{\theta}_{sem}$ is the corresponding sub-net parameters to be learned.
$\bm\gamma$ is optimized using the co-category labels as supervision.
Finally, we use the spatial and the semantic modulators as guidance to segment the co-object regions in each image $I^n$:
\begin{equation}
\label{eq:seg}
\textit{\textbf{M}}^n = f_{seg}(I^n,\textit{\textbf{S}}^n,\bm\gamma;\bm\theta_{seg}),
\end{equation}
where $\bm\theta_{seg}$ is the parameters of the segmentation sub-net.
To be specific, we transfer the spatial and semantic guidance $\{\mathcal{S},\bm\gamma\}$  into $f_{seg}$ using a simple shift-and-scale operation on the input CFMs of $f_{seg}$:
for each image $I^n\in \mathcal{I}$, its modulated feature maps are formulated as
\begin{equation}
\label{eq:modulator}
\textit{\textbf{Y}}^n_c = \gamma_c\textit{\textbf{X}}^n_c+\textit{\textbf{S}}^n, c=1,\ldots,d,
\end{equation}
where $\textit{\textbf{X}}^n_c$, $\textit{\textbf{Y}}^n_c\in \Re^{w\times h}$ are the input and output CFMs in the $c_{th}$ channel, $\gamma_c$ is the $c_{th}$ element of $\bm{\gamma}$.
\begin{figure}[t]
\centering
\includegraphics[width=0.4\textwidth]{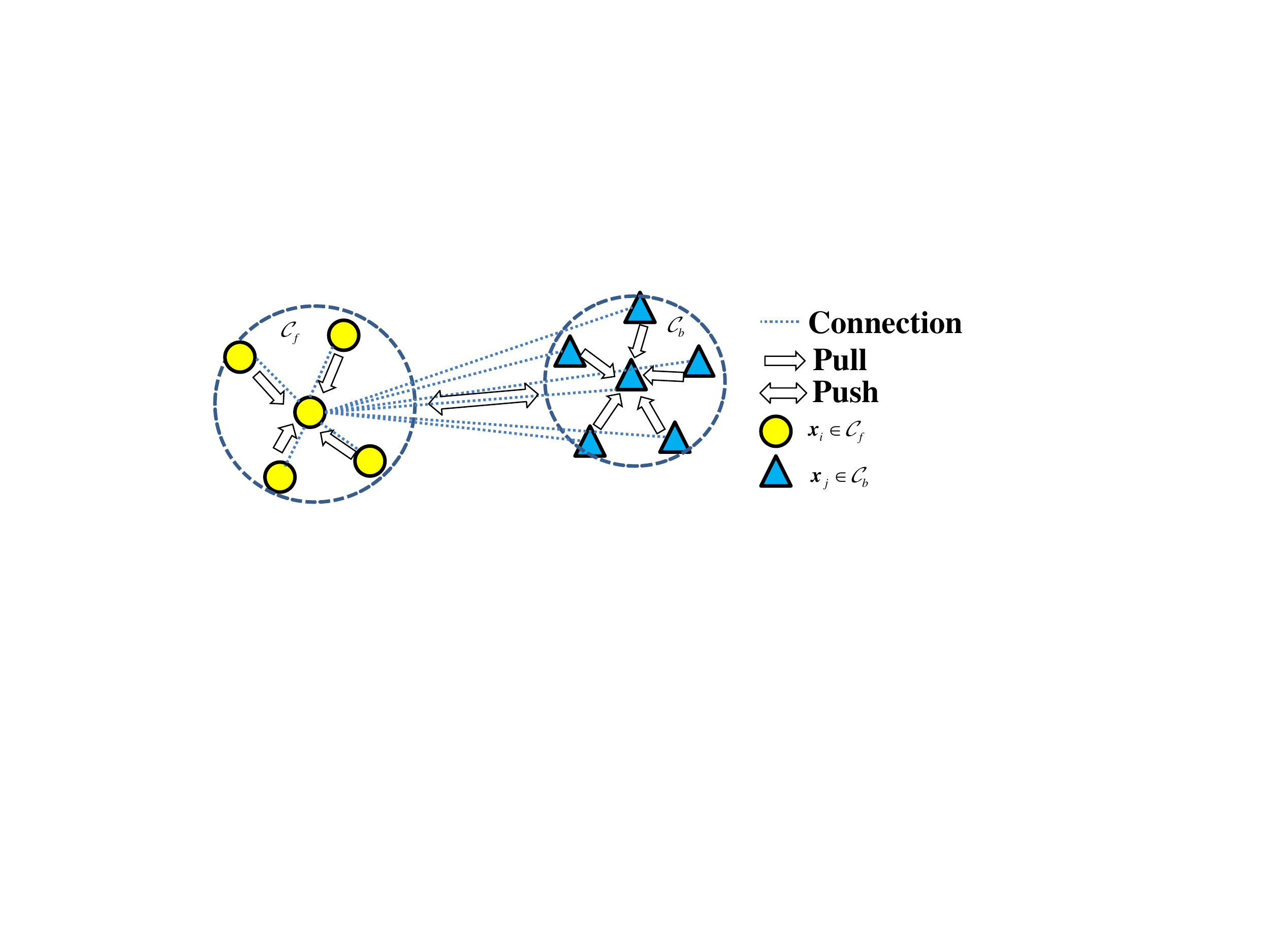} 
\caption{Schematic illustration of the proposed clustering objective for spatial modulator. The objective considers the mutual effects between any two samples, which pulls the samples of the same cluster together while pushing away the samples of different clusters. }
\label{fig:cluttering}
\end{figure}
\subsection{Spatial Modulator}
\hspace*{1em}
In the sub-net $f_{spa}$ (\ref{eq:spa modulator}),  the $i$-th channel feature $\textit{\textbf{x}}_i^n\in\Re^d$ of the input ${\bm\varphi}(I^n)\in\Re^{w\times h\times d}$ represents a corresponding local region in $I^n$.
For expression clarity, we represent all the channel feature representations of $\mathcal{I}$ as $\mathcal{X}=\{\textit{\textbf{x}}_i\in\Re^d\}_{i=1}^{whN}$.
The sub-net $f_{spa}$ aims at partitioning the data points in $\mathcal{X}$ into two classes $\mathcal{C}_f$, $\mathcal{C}_b$ of foreground and background.
However, if training $f_{spa}$ using a supervised learning method with a fixed set of categories, it cannot generalize well to unseen categories.
To this end, we propose a simple yet effective clustering approach to partitioning $\mathcal{X}$ into two clusters $\mathcal{C}_f$, $\mathcal{C}_b$ without knowing object semantic categories.
Our unsupervised method can highlight category-agnostic co-object regions in images and hence can better generalize to unseen categories.
As shown by Figure~\ref{fig:cluttering}, this can be achieved by maximizing all the distances between the foreground and the background
samples while minimizing all the distances between the foreground samples and between the background ones respectively. To this end, we define the clustering objective as follows:

\begin{equation}
\label{eq:spacluter}
\min\{\ell_{spa}=-2\sum_{i\in \mathcal{C}_f,j\in\mathcal{C}_b}d_{ij}+\sum_{i,j\in \mathcal{C}_f}d_{ij}+\sum_{i,j\in \mathcal{C}_b}d_{ij}\},
\end{equation}
where $d_{ij}=\|\textit{\textbf{x}}_i-\textit{\textbf{x}}_j\|_2^2$ is the squared Euclidean distance between samples $i$ and $j$.
Since we use normalized channel features satisfying $\|\textit{\textbf{x}}_i\|_2^2=1$, $d_{ij}$ can be reformulated as
\begin{equation}
\label{eq:dij}
d_{ij} = 2-2\textit{\textbf{x}}_i^\top \textit{\textbf{x}}_j.
\end{equation}
Using a cluster indictor vector $\textit{\textbf{s}}=[s_1,\ldots,s_{whN}]^\top$ subject to $\|\textit{\textbf{s}}\|_2^2=1$, where $s_i=1/\sqrt{whN}$ if $i\in \mathcal{C}_f$ and $s_i=-1/\sqrt{whN}$ if $i\in \mathcal{C}_b$, the loss function $\ell_{spa}$ in (\ref{eq:spacluter}) can be reformulated as
\begin{equation}
\label{eq:sparegularizor}
\ell_{spa}(\textit{\textbf{s}}) = whN\textit{\textbf{s}}^\top \textit{\textbf{D}}\textit{\textbf{s}},
\end{equation}
where the $(i,j)$-th entry of  $\textit{\textbf{D}}=d_{ij}$.
Putting (\ref{eq:dij}) into (\ref{eq:sparegularizor}) and removing the trivial constant $whN$, $\ell_{spa}$ can be reformulated as
\begin{equation}
\label{eq:spaloss}
\ell_{spa}(\textit{\textbf{s}}) = -\textit{\textbf{s}}^\top\textit{\textbf{G}}\textit{\textbf{s}},
\end{equation}
where $\textit{\textbf{G}}=\textit{\textbf{X}}^\top\textit{\textbf{X}}-\textit{\textbf{1}}$ with $\textit{\textbf{X}}=[\textit{\textbf{x}}_1,\ldots,\textit{\textbf{x}}_{whN}]$,
$\textit{\textbf{1}}$ denotes an all-ones matrix.
%
%
Relaxing the elements in $\textit{\textbf{s}}$ from binary indictor values to continuous values in $[-1,1]$ subject to $\|\textit{\textbf{s}}\|_2^2=1$, the solution $\widehat{\textit{\textbf{s}}}=\arg\min_{\textit{\textbf{s}}}\ell_{spa}(\textit{\textbf{s}})$ satisfies~\cite{Ding2004K}
\begin{equation}
\textit{\textbf{G}}\widehat{\textit{\textbf{s}}}=\lambda_{max}\widehat{\textit{\textbf{s}}},
\end{equation}
where $\lambda_{max}$ denotes the maximum eigenvalue of $\textit{\textbf{G}}$, and its corresponding eigenvector is $\widehat{\textit{\textbf{s}}}\in \Re^{whN}$.
The optimal solution $\widehat{\textit{\textbf{s}}}$ is then reshaped to a set of $N$ spatial masks $\{\widehat{\textit{\textbf{S}}}^n\in\Re^{w\times h}\}_{n=1}^N$ as the spatial guidance in (\ref{eq:modulator}).
\begin{figure}[t]
\centering
\includegraphics[width=0.88\columnwidth]{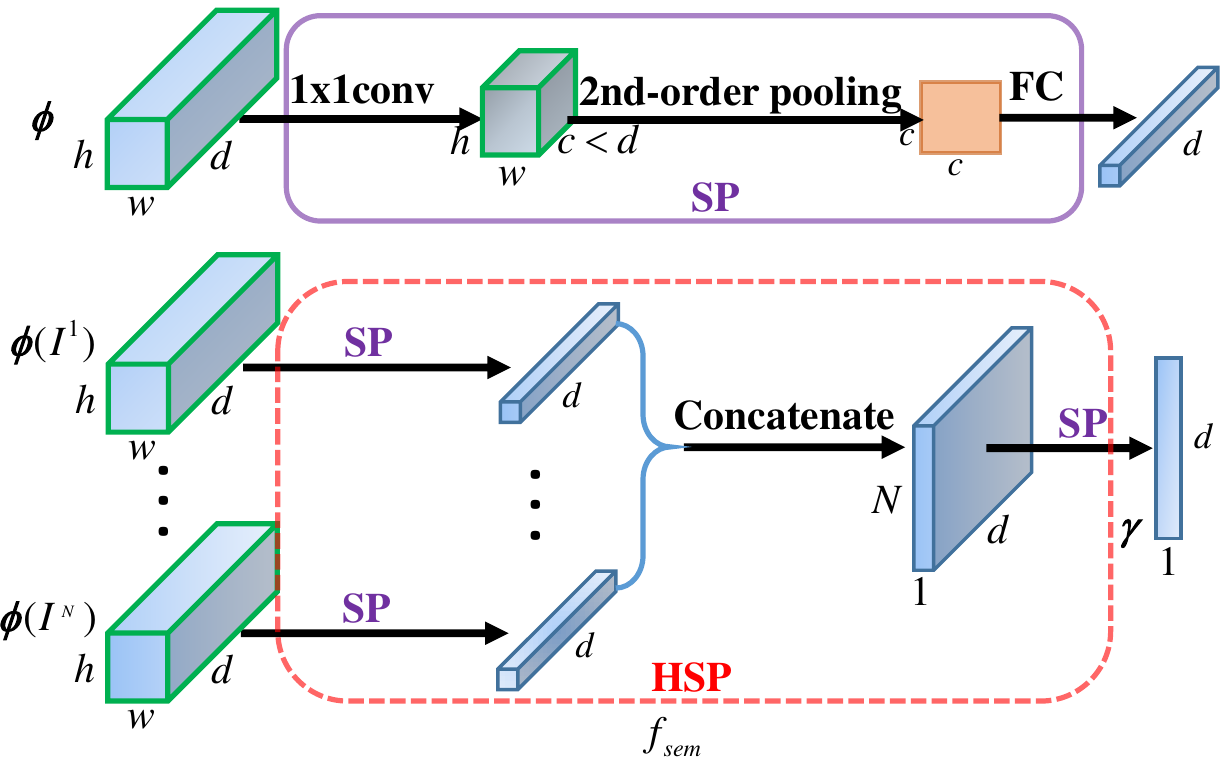} 
\caption{Illustration of the SP and the HSP. The sub-net $f_{sem}$ (\ref{eq:semantic modulator}) is composed of the HSP module.}
\label{fig:semanticmodulator}
\end{figure}
\subsection{Semantic Modulator}
\hspace*{1em}
Figure~\ref{fig:semanticmodulator} shows the diagram of the key modules in the sub-net $f_{sem}$ (\ref{eq:semantic modulator}), including the SP and the HSP.
The SP exploits the high-order statistics of the holistic representation to enhance the non-linear representative capability of the learned model~\cite{gao2019global}, while the HSP can capture the long-range dependency along channel dimension of the group-wise feature tensors, paying more attention to important channels for classification task under the supervision of co-category labels.
%
%
%

\textbf{SP:} Given input feature tensor $\bm\phi\in \Re^{w\times h\times d}$,  we firstly leverage a $1\times 1$ convolution to reduce the number of channels from $d$ to $c$ to reduce the computational cost for the following operations.
Then, we compute pairwise channel correlations of the $w\times h\times c$ tensor to yield a $c \times c$ covariance matrix.
Each entry in the $c \times c$ covariance matrix measures the relevance between the feature maps in two channels, which leverages a quadratic operator to model high-order statistics of the holistic representation, hence enabling to enhance non-linear modeling capability.
Afterwards, we use an FC layer to transform the $c\times c$ covariance matrix into a $1\times 1\times d$ tensor that indicates the feature channel importance.

\textbf{HSP:} For each image $I^n\in \mathcal{I}$, its feature tensor $\bm\phi(I^n)$ is fed into an SP layer, outputting a $1\times 1\times d$ indicator tensor.
Then, all the indictor tensors are concatenated vertically to yield a group-wise semantic representation, which is again fed into an SP layer to capture the long-range dependency along the channel dimension of the group-wise semantic representation, yielding an indictor vector $\bm\gamma$ that steers attention to the important channels that are essential for co-category classification.

\textbf{Loss:} The output $\bm\gamma$ of $f_{sem}$ is followed by an FC layer and a sigmoid layer, yielding a classifier response:
 \begin{equation}
 \label{eq:cls}
 \widehat{\textit{\textbf{y}}}=sigmoid(\textit{\textbf{W}}\bm\gamma+\textit{\textbf{b}}),
 \end{equation}
 where $\textit{\textbf{W}}\in\Re^{L\times d}$ and $\textit{\textbf{b}}\in\Re^{L}$ are the parameters of the FC layer, $L$ denotes the number of the co-category in the training set.

 The widely used cross-entropy loss function for classification is adopted to learn the indictor $\bm\gamma$ in (\ref{eq:cls}):
\begin{equation}
\label{eq:semloss}
\ell_{sem} = - \frac{1}{L} \sum\limits_{l=1}^{L} y_{l} \log \widehat{y}_{l} - (1 - y_{l}) \log (1 - \widehat{y}_{l}),
\end{equation}
where $\widehat{y}_{l}$ is the $l$-th entry of $\widehat{\textit{\textbf{y}}}$ that is the prediction value for the $l$-th co-category and $y_l\in\{0,1\}$ is the ground-truth label.
%
%

%

\subsection{Segmentation Sub-net}
\hspace*{1em}
Given the input group-wise CFMs $\{\textit{\textbf{X}}^n\}_{n=1}^N$ of the images $\mathcal{I}$, the sub-net $f_{seg}$ are modulated by the outputs $\{\mathcal{S},\bm\gamma\}$ of $f_{spa}$ and $f_{sem}$, yielding a group of modulated representations $\{\textit{\textbf{Y}}^n\}_{n=1}^N$ using (\ref{eq:modulator}).
Each $\textit{\textbf{Y}}^n$ is composed of a group of multi-resolution representations $\{\textit{\textbf{R}}_i^n\}_{i=1}^4$. Similar to the FPN~\cite{Lin2017Feature}, we fuse $\{\textit{\textbf{R}}_i^n\}_{i=1}^4$ from coarse to fine: with the coarser-resolution feature maps, we use a $1\times 1$ convolution layer to make the channel number equal to the corresponding top-down ones, following by an upsampling layer to make their spatial resolutions the same. Then, the upsampled maps are merged with the corresponding top-down ones via element-wise addition.
The process is repeated until the finest resolution maps are generated as $\textit{\textbf{R}}^n=\textit{\textbf{R}}_1^n\oplus\cdots\oplus\textit{\textbf{R}}_{4}^n$. Finally, the maps $\textit{\textbf{R}}^n$ are fed into a convolutional layer, following by a $1\times 1$ convolutional layer and an upsampling layer to generate the corresponding segmentation mask $\widehat{\textit{\textbf{M}}}^n$.
Denoting the ground-truth binary co-segmentation masks in the training image group as $\mathcal{M}_{gt} =\{\textit{\textbf{M}}_{gt}^n\}_{n=1}^{N}$, the loss function for the segmentation task is formulated as a weighted cross-entropy loss for pixel-wise classification:
%
\begin{align}
\label{eq:segloss}
\ell_{seg} =&- \frac{1}{NP} \sum_{n=1}^N\sum_{i=1}^{P}\{\delta^n \textit{\textbf{M}}_{gt}^{n}(i)\log \widehat{\textit{\textbf{M}}}^{n}(i)\notag\\&- (1-\delta^n)(1-\textit{\textbf{M}}_{gt}^{n}(i))\log (1-\widehat{\textit{\textbf{M}}}^{n}(i))\},
\end{align}
%
where $P$ is the number of the pixels in each training image, $i$ denotes the pixel index,
$\delta^n$ is the ratio between all positive pixels and all pixels in image $I^n$, which balances the positive and negative samples.
%
%
\subsection{Loss Function}
\hspace*{1em}
The three sub-nets $f_{spa}$, $f_{sem}$ and $f_{seg}$ are trained jointly via optimizing the following multi-task loss function
\begin{equation}
\label{eq:loss}
\ell = \ell_{spa}+\ell_{sem}+\ell_{seg},
\end{equation}
where $\ell_{spa}$, $\ell_{sem}$ and $\ell_{seg}$ are defined by (\ref{eq:spaloss}) , (\ref{eq:semloss}) and (\ref{eq:segloss}), respectively.
\section{Experiments}
\label{sec:experiments}
%
%
\subsection{Implementation Details}
\hspace*{1em}
We leverage the HRNet \cite{sun2019deep} pre-trained on ImageNet \cite{deng2009imagenet} as the backbone network to extract the multi-resolution semantic features.
Moreover, we also report the results of using the VGG16 backbone network~\cite{simonyan2014very}, which still demonstrate competing performance over state-of-the-art methods.
Except for using the pretrained backbone network parameters as initialization, all other parameters are trained from scratch.
We follow the same settings as  \cite{wei2017group,wang2019robust}: the input image group $\mathcal{I}$ consists of $N=5$ images that are randomly selected from a group of images with co-object category, and a mini-batch of $4\times \mathcal{I}$ is fed into the model simultaneously during training.
All images in $\mathcal{I}$ are resized to $224\times 224$ as input, and then the predicted co-segmentation maps are resized to the original image sizes as outputs.
We leverage the Adam algorithm \cite{kingma2014adam} to optimize the whole network in an end-to-end manner, among which the exponential decay rates for estimating the first and the second moments are set to $0.9$ and $0.999$, respectively.
The learning rate starts from 1e-4 and reduces by a half every $25,000$ steps until the model converges at about 200,000 steps.
%
%
Our model is implemented in PyTorch and a Nvidia RTX $2080$Ti GPU is adopted for acceleration.
We adopt the COCO-SEG dataset released by~\cite{wang2019robust} to train our model.
The dataset contains $200,000$ images belonging to $L=78$ groups, among which each image has a manually labeled binary mask with co-category label information.
The training process takes about $40$ hours.
\subsection{Datasets and Evaluation Metrics}
%

~~\textbf{Datasets:} We conduct extensive evaluations on four widely-used benchmark datasets~\cite{faktor2013co,rubinstein2013unsupervised} including sub-set of MSRC, Internet, sub-set of iCoseg, and PASCAL-VOC. Among them, the sub-set of MSRC includes $7$ classes: bird, car, cat, cow, dog, plane, sheep, and each class contains $10$ images.
The Internet has $3$ categories of airplane, car and horse. Each class has $100$ images including some images with noisy labels.
The sub-set of iCoseg contains $8$ categories, and each has a different number of images.
The PASCAL-VOC is the most challenging dataset with $1,037$ images of $20$ categories selected from the PASCAL-VOC 2010 dataset~\cite{Everingham10}.

\textbf{Evaluation Metrics:} We adopt two widely-used metrics to evaluate the co-segmentation results, including the \textit{precision} $\mathcal{P}$ and the \textit{Jaccard index} $\mathcal{J} $.
The precision $\mathcal{P}$ measures the percentage of the correctly segmented pixels for both foreground and background, while the Jaccard index $\mathcal{J}$ is defined as the intersection area of the predicted foreground objects and the ground truth divided by their union area.
\renewcommand\arraystretch{1.2}
\begin{table}[t]
\caption{Quantitative comparison results on the sub-set of MSRC. The bold numbers indicate the best results. }\smallskip
\centering
\resizebox{.95\columnwidth}{!}{
\smallskip\begin{tabular}{|l||c|c|}
\hline
\multicolumn{1}{|c||}{MSRC} & Ave. $\mathcal{P} ($\%$)$ &Ave. $\mathcal{J}$ ($\%$)\\
\hline
\cite{vicente2011object} & 90.2 & 70.6 \\
\cite{rubinstein2013unsupervised} & 92.2 & 74.7 \\
\cite{wang2013image} & 92.2 & - \\
\cite{faktor2013co} & 92.0 & 77.0 \\
\cite{mukherjee2018object} & 84.0 & 67.0 \\
\cite{li2018deep} & 92.4 & 79.9 \\
\cite{chen2018semantic} & \textbf{95.2} & 77.7\\
\hline\hline
Ours-VGG16 & 94.3 & 79.4 \\
Ours-HRNet & \textbf{95.2} & \textbf{81.9} \\
\hline
\end{tabular}
}
\label{MSRC}
\end{table}
\renewcommand\arraystretch{1.2}
\begin{table}[t]
\caption{Quantitative comparison results on the Internet. The bold numbers indicate the best results.  }\smallskip
\centering
\resizebox{1\columnwidth}{!}{
\smallskip\begin{tabular}{|l||cc|cc|cc|}
\hline
\multirow{2}{*}{~~~~~~~~~~~~~~~~~~~~~Internet}
 & \multicolumn{2}{|c|}{Airplane} & \multicolumn{2}{c|}{Car} & \multicolumn{2}{c|}{Horse} \\
 & Ave. $\mathcal{P}$ ($\%$) & Ave. $\mathcal{J}$ ($\%$) & Ave. $\mathcal{P}$ ($\%$) & Ave. $\mathcal{J}$ ($\%$) & Ave. $\mathcal{P}$ ($\%$) & Ave. $\mathcal{J}$ ($\%$) \\
\hline
\cite{joulin2012multi} & 47.5 & 11.7 & 59.2 & 35.2 & 64.2 & 29.5\\
\cite{rubinstein2013unsupervised} & 88.0 & 55.8 & 85.4 & 64.4 & 82.8 & 51.6\\
\cite{chen2014enriching} & 90.2 & 40.3 & 87.6 & 64.9 & 86.2 & 33.4\\
\cite{jerripothula2016image} & 90.5 & 61.0 & 88.0 & 71.0 & 88.3 & 60.0\\
\cite{quan2016object} & 91.0 & 56.3 & 88.5 & 66.8 & 89.3 & 58.1\\
\cite{sun2016learning} & 88.6 & 36.3 & 87.0 & 73.4 & 87.6 & 54.7\\
\cite{tao2017image} & 79.8 & 42.8 & 84.8 & 66.4 & 85.7 & 55.3\\
\cite{yuan2017deep} & 92.6 & 66.0 & 90.4 & 72.0 & 90.2 & 65.0\\
\cite{li2018deep} & 94.1 & 65.4 & 93.9 & \textbf{82.8} & 92.4 & 69.4\\
\cite{chen2018semantic} & - & 65.9 & - & 76.9 & - & 69.1\\
\cite{MaCoSNet} & 94.1 & 65.0 & \textbf{94.0} & 82.0 & 92.2 & 63.0\\
\hline\hline
Ours-VGG16 & 94.6 & 66.7 & 89.7 & 68.1 & 93.2 & 66.2\\
Ours-HRNet & \textbf{94.8} & \textbf{69.6} & 91.6 & 82.5 & \textbf{94.4} & \textbf{70.2}\\
\hline
\end{tabular}
}
\label{Internet}
\end{table}
\renewcommand\arraystretch{1.3}
\begin{table*}[t]
\caption{Quantitative comparison results on the sub-set of iCoseg. The bold numbers indicate the best results.}\smallskip
\centering
\resizebox{2.0\columnwidth}{!}{
\smallskip\begin{tabular}{|l||c|cccccccc|}
\hline
\multicolumn{1}{|c||}{iCoseg} & Ave. $\mathcal{J}$ ($\%$) & bear2 & brownbear & cheetah & elephant & helicopter & hotballoon & panda1 & panda2 \\
\hline
\cite{rubinstein2013unsupervised} & 70.2 & 65.3 & 73.6 & 69.7 & 68.8 & 80.3 & 65.7 & 75.9 & 62.5\\
\cite{jerripothula2014automatic} & 73.8 & 70.1 & 66.2 & 75.4 & 73.5 & 76.6 & 76.3 & 80.6 & 71.8\\
\cite{faktor2013co} & 78.2 & 72.0 & 92.0 & 67.0 & 67.0 & \textbf{82.0} & 88.0 & 70.0 & 55.0\\
\cite{jerripothula2016image} & 70.4 & 67.5 & 72.5 & 78.0 & 79.9 & 80.0 & 80.2 & 72.2 & 61.4\\
\cite{li2018deep} & 84.2 & 88.3 & \textbf{92.0} & 68.8 & 84.6 & 79.0 & 91.7 & 82.6 & 86.7\\
\cite{chen2018semantic} & 86.0 & 88.3 & 91.5 & 71.3 & 84.4 & 76.5 & 94.0 & \textbf{91.8} & \textbf{90.3}\\
\hline\hline
Ours-VGG16 & 88.0 & 87.4 & 90.3 & 84.9 & 90.6 & 76.6 & 94.1 & 90.6 & 87.5\\
Ours-HRNet & \textbf{89.2} & \textbf{91.1} & 89.6 & \textbf{88.6} & \textbf{90.9} & 76.4 & \textbf{94.2}& 90.4 & 87.5\\
\hline
\end{tabular}
}
\label{iCoseg}
\end{table*}
\renewcommand\arraystretch{1.3}
\begin{table*}[t]
\caption{Quantitative comparison results on the PASCAL-VOC. The bold numbers indicate the best results.}\smallskip
\centering
\resizebox{2.1\columnwidth}{!}{
\smallskip\begin{tabular}{|l||c|c|cccccccccccccccccccc|}
\hline
\multicolumn{1}{|c||}{PASCAL-VOC} & Ave. $\mathcal{P}$ ($\%$) & Ave. $\mathcal{J}$ ($\%$) & A.P. & Bike & Bird & Boat & Bottle & Bus & Car & Cat & Chair & Cow & D.T. & Dog & Horse & M.B. & P.S. & P.P. & Sheep & Sofa & Train & TV\\
\hline
\cite{faktor2013co} & 84.0 & 46 & 65 & 14 & 49 & 47 & 44 & 61 & 55 & 49 & 20 & 59 & 22 & 39 & 52 & 51 & 31 & 27 & 51 & 32 & 55 & 35\\
\cite{lee2015multiple} & 69.8 & 33 & 50 & 15 & 29 & 37 & 27 & 55 & 35 & 34 & 13 & 40 & 10 & 37 & 49 & 44 & 24 & 21 & 51 & 30 & 42 & 16\\
\cite{chang2015optimizing} &82.4 & 29 & 48 & 9 & 32 & 32 & 21 & 34 & 42 & 35 & 13 & 50 & 6 & 22 & 37 & 39 & 19 & 17 & 41 & 21 & 41 & 18\\
\cite{quan2016object} & 89.0 & 52 & - & - & - & - & - & - & - & - & - & - & - & - & - & - & - & - & - & - & - & - \\
\cite{hati2016image} & 72.5 & 25 & 44 & 13 & 26 & 31 & 28 & 33 & 26 & 29 & 14 & 24 & 11 & 27 & 23 & 22 & 18 & 17 & 33 & 27 & 26 & 25\\
\cite{jerripothula2016image} & 85.2 & 45 & 64 & 20 & 54 & 48 & 42 & 64 & 55 & 57 & 21 & 61 & 19 & 49 & 57 & 50 & 34 & 28 & 53 & 39 & 56 & 38\\
\cite{jerripothula2017object} & 80.1 & 40 & 53 & 14 & 47 & 43 & 42 & 62 & 50 & 49 & 20 & 56 & 13 & 38 & 50 & 45 & 29 & 26 & 40 & 37 & 51 & 37\\
\cite{wang2017multiple} & 84.3 & 52 & 75 & 26 & 53 & 59 & 51 & 70 & 59 & 70 & 35 & 63 & 26 & 56 & 63 & 59 & 35 & 28 & 67 & 52 & 52 & 48\\
\cite{li2018deep} & 94.2 & 65 & - & - & - & - & - & - & - & - & - & - & - & - & - & - & - & - & - & - & - & - \\
\cite{hsu2018co} & 91.0 & 60 & 77 & 27 & 70 & 61 & 58 & 79 & 76 & 79 & 29 & 75 & \textbf{28} & 63 & 66 & 65 & 37 & 42 & 75 & 67 & 68 & 51 \\
\hline\hline
Ours-VGG16 & 93.7 & 66 & \textbf{83} & 35 & \textbf{75} & 69 & 58 & 87 & 77 & \textbf{80} & 26 & 86 & 7 & 74 & 79 & 71 & 45 & 39 & 81 & 68 & 83 & 59 \\
Ours-HRNet & \textbf{94.9} & \textbf{71} & 82 & \textbf{37} & 74 & \textbf{70} & \textbf{67}& \textbf{88} & \textbf{82} & 77 & \textbf{36} & \textbf{87} & 15 & \textbf{75} & \textbf{82} & \textbf{72} & \textbf{58} & \textbf{46} & \textbf{82} & \textbf{77} & \textbf{84} & \textbf{69}\\
\hline
\end{tabular}
}
\label{PASCALVOC}
\end{table*}
\subsection{Results}
\hspace*{1em}
We quantitatively and qualitatively compare our algorithm with several state-of-the-art co-segmentation methods on the four benchmark datasets.

\textbf{Quantitative Results:}
Tables \ref{MSRC}, \ref{Internet}, \ref{iCoseg}, \ref{PASCALVOC} list the comparison results of  our method with other state-of-the-arts on the sub-set of MSRC, Internet, sub-set of iCoseg and PASCAL-VOC.
For fair comparisons, the reported results of the compared methods are directly obtained from their publications.
We can observe that our algorithm outperforms the other state-of-the-arts in term of both metrics on most object categories in each dataset.
Especially on the PASCAL-VOC, which has more challenging scenarios, the proposed algorithm achieves the best average $\mathcal{P}$ and average $\mathcal{J}$  with a score of $94.9\%$ and $71\%$, respectively, significantly outperforming the others by a large margin.
Moreover, on the sub-set of MSRC and sub-set of iCoseg, the average $\mathcal{J}$ by our method has a score of $81.9\%$ and $89.2\%$, outperforming the others by about $3\%$.
Besides, on the Internet, our algorithm achieves the best performance on airplane and horse categories, as well as a competitive performance on car category in terms of both metrics average $\mathcal{P}$ and average $\mathcal{J}$.
\begin{figure*}[t]
\centering
\includegraphics[width=0.497\textwidth]{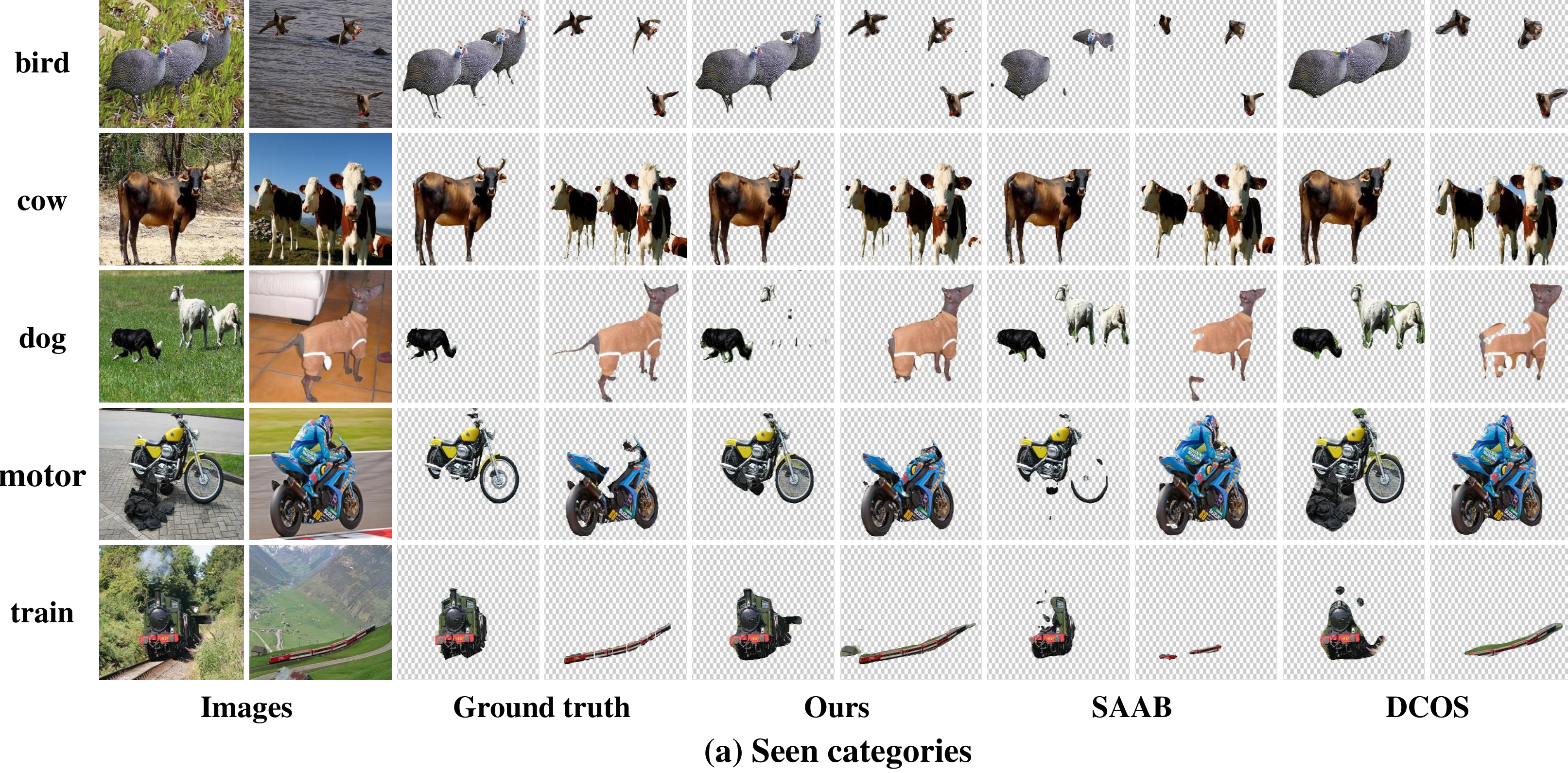} 
\includegraphics[width=0.497\textwidth]{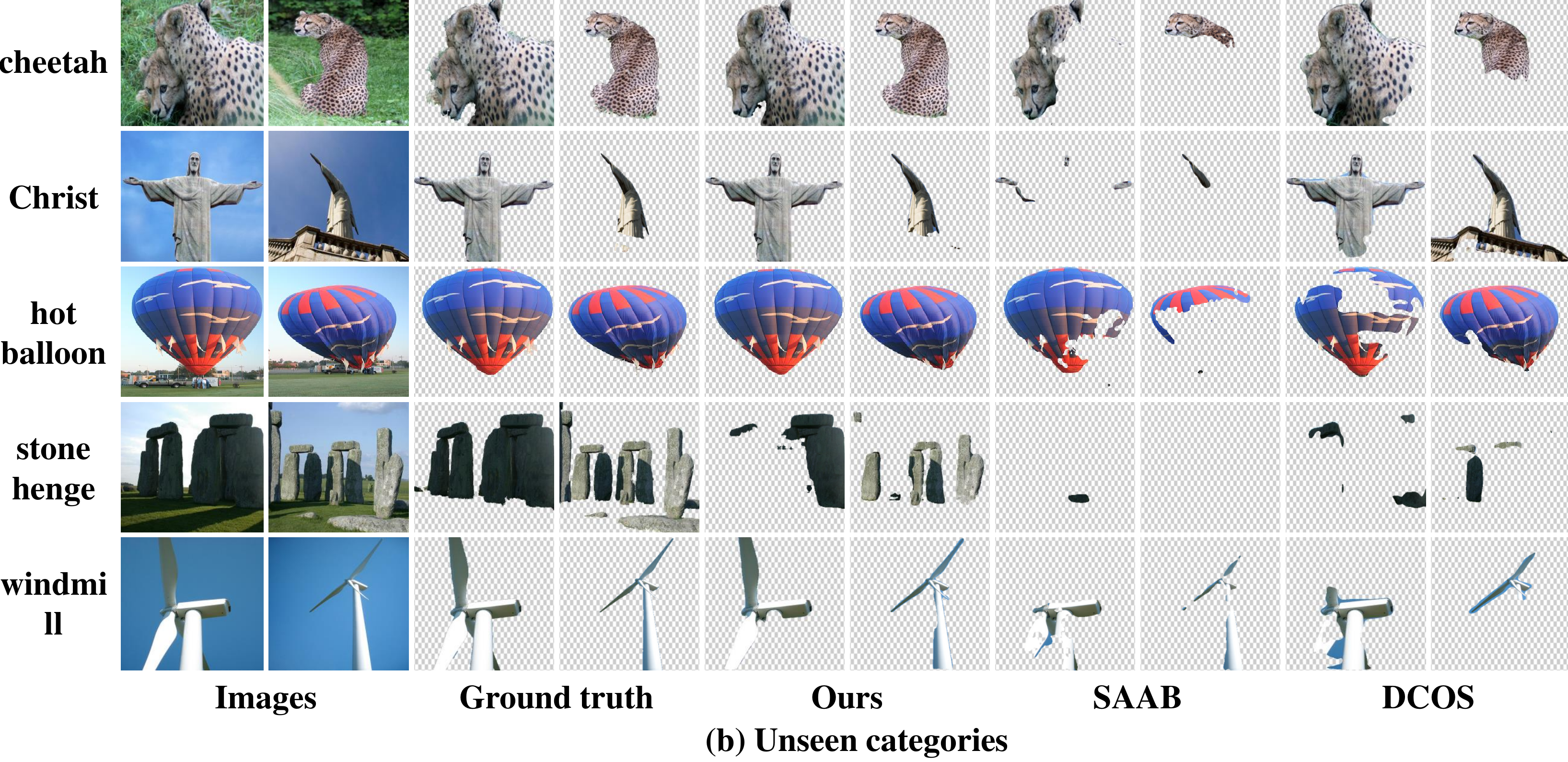}
\caption{Some qualitative comparison results generated by the proposed method, SAAB~\cite{chen2018semantic} and DCOS~\cite{li2018deep} for co-segmenting objects associated to the training categories and unseen categories, respectively.}
\label{coseg}
\end{figure*}

\textbf{Qualitative Results:}
Figure \ref{coseg} shows some qualitative results by comparing our method with SAAB~\cite{chen2018semantic} and DCOS~\cite{li2018deep}.
Those images are chosen from all of the four datasets composed of co-objects with seen categories (inside the $78$ categories of the COCO-SEG) and unseen categories (outside the categories of the COCO-SEG).
For the seen categories shown by Figure \ref{coseg}(a), we can observe that SAAB and DCOS cannot discover the co-objects in the dog group accurately and two distractors (sheep) have been mis-classified as co-objects.
However, the proposed approach does not suffer from this issue since it uses co-category labels as supervision to learn an effective semantic modulator that can well capture high-level semantic category information.
Besides, as shown by Figure~\ref{coseg}(a), (b), the proposed approach can discover the whole co-objects of seen and unseen categories well because its spatial modulator is learned by an unsupervised method that can not only help to locate the co-object regions of seen categories well, but also generalize well to unseen categories.
\renewcommand\arraystretch{1.2}
\begin{table}[t]
\caption{Ablative experiments of the proposed model on the PASCAL-VOC. The bold numbers indicate the best results. The symbol `$-f$' denotes removing the module $f$.}\smallskip
\centering
\resizebox{.75\columnwidth}{!}{
\smallskip\begin{tabular}{|l||cc|}
\hline
\multicolumn{1}{|c||}{PASCAL-VOC} & Avg. $\mathcal{P}$ ($\%$) & Avg. $\mathcal{J}$ ($\%$)\\
\hline
$f_{spa}\&f_{sem}\&f_{seg}$ & \textbf{94.9} & \textbf{71} \\
$-$ $f_{spa}$  & 94.5 & 69 \\
$-$ $f_{sem}$ & 85.0 & 38  \\
$-$ ($f_{spa}\&f_{sem}$) & 82.0 & 27  \\
\hline
\end{tabular}
}
\label{fig:ablative}
\end{table}
\subsection{Ablative Study }
\hspace*{1em}
To further show our main contributions, we compare different variants of our model including those without spatial modulator ($-f_{spa}$), semantic modulator ($-f_{sem}$) and both modulators ($-(f_{spa}\&f_{sem})$), respectively.
Table~\ref{fig:ablative} lists the results of ablative experiments on the PASCAL-VOC.
We can observe that without $f_{spa}$, the average $\mathcal{P}$ score drops from $94.9\%$ to $94.5\%$ while the average $\mathcal{J}$ score reduces by $2\%$ from $71\%$ to $69\%$, which verifies the effectiveness of the proposed module $f_{spa}$.
Moreover, without $f_{sem}$, the performance suffers from a significant loss with a big drop of $9.9\%$ and $33\%$ for the  average $\mathcal{P}$ and $\mathcal{J}$ scores, respectively, indicating the critical role of the semantic modulator as a guidance to learn an effective segmentation network for accurate co-segmentation.
Besides, compared to that only removes $f_{sem}$, removing both modulators $f_{spa}$ and $f_{sem}$ further makes the performance of our model drop by $3\%$ and $11\%$ in terms of average $\mathcal{P}$ and average $\mathcal{J}$, respectively. These experiments confidently validate that both modulators have a positive effect to boost the performance of our model.
\section{Conclusions}
\label{sec:conclusions}
\hspace*{1em}
In this paper, we have presented a spatial-semantic modulated deep network framework for object co-segmentation. Our model is composed of a spatial modulator, a semantic modulator and a segmentation sub-net.
The spatial modulator is to learn a mask to coarsely localize the co-object regions in each image that captures the correlations of image feature descriptors with unsupervised learning.
The semantic modulator is to learn a channel importance indictor under the supervision of co-category labels. We have proposed the HSP module to transform the input image features of the semantic modulator for classification use.
The outputs of the two modulators manipulate the input feature maps of the segmentation sub-net by a simple shift-and-scale operation to adapt it to target on segmenting the co-object regions.
Both quantitative and qualitative evaluations on four image co-segmentation benchmark datasets have demonstrated superiority of the proposed method to the state-of-the-arts.
\section*{Acknowledgments}
This work is supported in part by National Major Project of China for New Generation of AI (No. 2018AAA0100400), in part by the Natural
Science Foundation of China under Grant nos. 61876088, 61825601, in part by the Natural Science Foundation of Jiangsu Province under Grant no. BK20170040.

\bibliography{ref}
\bibliographystyle{aaai}

\end{document}